\documentclass{article}

\usepackage[english]{babel}

\usepackage[letterpaper,top=2cm,bottom=2cm,left=3cm,right=3cm,marginparwidth=1.75cm]{geometry}

\usepackage{amsmath}
\usepackage{graphicx}
\usepackage[colorlinks=true, allcolors=blue]{hyperref}
\usepackage{algorithm}
\usepackage{algorithmic}
\usepackage{subcaption}

\usepackage{amsfonts}

\title{Parallel Multi-Scale Networks with Deep Supervision for Hand Keypoint Detection}
\author{Renjie Li, Son Tran, Saurabh Garg, Katherine Lawler, Jane Alty, Quan Bai}

\begin{document}
\maketitle

\begin{abstract}
Keypoint detection plays an important role in a wide range of applications. However, predicting keypoints of small objects such as human hands is a challenging problem. Recent works fuse feature maps of deep Convolutional Neural Networks (CNNs), either via multi-level feature integration or multi-resolution aggregation. Despite achieving some success, the feature fusion approaches increase the complexity and the opacity of CNNs. To address this issue, we propose a novel CNN model named Multi-Scale Deep Supervision Network (P-MSDSNet) that learns feature maps at different scales with deep supervisions to produce attention maps for adaptive feature propagation from layers to layers. P-MSDSNet has a multi-stage architecture which makes it scalable while its deep supervision with spatial attention improves transparency to the feature learning at each stage. We show that P-MSDSNet outperforms the state-of-the-art approaches on benchmark datasets while requiring fewer number of parameters. We also show the application of P-MSDSNet to quantify finger tapping hand movements in a neuroscience study.
\end{abstract}

\section{Introduction}

Hand keypoint detection is a method that identifies the position of different keypoints in images of hands. The keypoints of a hand are the fingertips and joints on the fingers such as knuckles. Hand keypoint detection methods have largely been used to extract hand movement features~\cite{dardas2011real,zhang2020mediapipe}. Such movement features can then be used in different domains such as healthcare and human-computer interaction~\cite{rautaray2015vision}.  
Recent approaches for keypoint detection across a range of areas rely heavily on Convolutional Neural Networks (CNNs) thanks to their ability in learning discriminable features from visual data~\cite{lecun1995convolutional}. By stacking multiple blocks of convolutional layers one on top of another, along with other operators such as pooling and normalisation, CNNs can learn different levels of abstractions of visual features that may improve the effectiveness of localising the positions of interest \cite{huang2015deepfinger,wu2017yolse, yang2020msb}. However, the different sizes of objects in images pose a great challenge for learning reliable features for the prediction tasks. Several architectures have been proposed to deal with the issue, including Hourglass model~\cite{newell2016stacked}, HigherHRNet~\cite{cheng2020higherhrnet}, U-Net~\cite{ronneberger2015u} and Fully Convolutional Networks (FCN)~\cite{long2015fully}. These approaches aim to learn and fuse features at different scales to improve the performance by connecting multi-resolution sub-networks. In \cite{yang2020msb}, multi-scale features were extracted from a large block of feature maps, which were then fine-tuned and aggregated for prediction. Despite achieving some success, the feature  fusion  approaches  increase  the  complexity  and  the opacity of CNNs. For hand keypoint detection, in particular, such scale variance problem and the lack of transparency in deep architecture still present major obstacles to the employment of CNNs in real-world applications.

To address the aforementioned limitations, we propose Multi-Scale Deep Supervision Network (P-MSDSNet), a novel deep neural network architecture with deeply supervised attention at multiple scales. The network is constructed in a modular fashion by stacking multiple stages, one on top of another. All stages have the same structure, each consisting of a set of multi-resolution sub-networks connected in parallel. The sub-networks are initialised by convolutional operators with different strides to learn features at different scales. This idea aims to reduce the need for another pooling layer as in \cite{zhang2017amulet}. We extend the idea of deep supervision \cite{lee2015deeply, li2018deep} to learn attention maps for propagating the feature maps from one stage to another. With this process, the model can guide the flow of information to focus on spatial features that represent the keypoints. Although the deeply supervised attention does not provide a complete level of transparency, it offers a means to monitor the behavior of the deep network, i.e. what is learned at each stage and what information is forwarded to the next stage.

We conducted experiments to evaluate the effectiveness of P-MSDSNet on hand keypoint benchmark datasets. The results show that P-MSDSNet outperforms state-of-the-art deep learning models in hand keypoint detection while requiring fewer parameters. Furthermore, we demonstrate that P-MSDSNet can self-tune the attention maps in a multi-stage architecture. Finally, we show a real-world application of P-MSDSNet in estimating finger tapping movements for a neuroscience study.

Our contributions are summarised as follows:
\begin{itemize}
\item[$\bullet$] A flexible architecture to effectively learn and fuse different context information at different scales and depths.
\item[$\bullet$] A deeply supervised attention module to guide the learning and control the propagation of information in deep models.
\item[$\bullet$] An effective and more transparent solution for real-world applications.
\end{itemize}

\section{Related Work}  
\subsection{Hand Keypoint Detection}
Early works on hand keypoint detection focused on using features that were manually extracted from the image by humans, such as hand direction~\cite{grzejszczak2016hand} and color features~\cite{raheja2012fingertip}. The performances of these works were limited by uncontrolled environments and self-occlusions~\cite{erol2007vision}. Recently, CNNs based deep learning methods have proved to be an effective technique to address the hand keypoint detection problem. In \cite{wu2017yolse}, CNNs were used to integrate features at different depth levels to regress hand fingertip positions. Predicting hand keypoints from images is challenging as there can be redundant features such as human body parts which may interfere the performance. One approach to mitigate such impact is to focus the learning on the hand area. For example, in \cite{huang2015deepfinger} the authors applied a cascade of CNNs to detect finger keypoints. The first CNN was used to obtain hand area and the second CNN was used to regress the positions of hand fingertips. In \cite{mukherjee2019fingertip}, Faster-RCNN \cite{ren2016faster} was used to localise hands in images and the hand fingertips were detected by calculating the distances between the center and contour points as well as the curvature entropy at each contour point.

In this paper, we employ spatial attention to guide the model to focus on the potential areas of the keypoints. Therefore, our approach only needs one step of prediction and reduces the annotation cost.

\subsection{Multi-Scale Networks with Auxiliary Supervision}
Multi-scale networks have been designed to learn invariant features from input images. Recent approaches learn and fuse the features through down-up scaling processes (also known as high-to-low and low-to-high processes). U-Net~\cite{ronneberger2015u} and  Fully Convolutional Networks (FCN)~\cite{long2015fully} adopted an encoder-decoder paradigm to extract small and large scale features for image segmentation. Amulet~\cite{zhang2017amulet} and Bidirectional Fully Connected Network (MSBFCN)~\cite{yang2020msb} extracted and fused multi-scale features from pre-defined models, i.e. VGG-16 \cite{simonyan2014very} and ResNet \cite{he2016deep} respectively. The Stacked Hourglass network~\cite{newell2016stacked} applied encoder-decoder architecture and skip connections to fuse different scale and depth information. HRNet \cite{sun2019deep} and HigherHRNet~\cite{cheng2020higherhrnet} connected multi-scale sub-networks in parallel and aggregated the feature maps of the sub-networks to maintain high-resolution representations. 

In deep learning,  multi-scale feature fusion is usually coupled with auxiliary supervision (also known as deep supervision/intermediate supervision), adding supervision at different stages of neural networks \cite{li2018deep}. The key advantage of auxiliary supervision is to improve the learning of discriminable features for prediction tasks \cite{zhang2018deep,ronneberger2015u,yang2020msb}. This technique has been applied largely for  keypoint detection \cite{li2018deep, newell2016stacked, wei2016convolutional}.

The P-MSDSNet architecture is inspired by `down-up' scaling idea in \cite{sun2019deep,cheng2020higherhrnet} but is different in that P-MSDSNet maintains both low-resolution and high-resolution representations in parallel sub-networks. For auxiliary supervision, instead of learning discriminable features explicitly as in \cite{lee2015deeply}, it learns a spatial attention map at each stage. The advantage of this idea is twofold. First, P-MSDSNet can focus on the discriminable features in the areas surrounding the keypoints. Second, the attention map can help shed light on the learning at intermediate layers, thus offering a certain degree of transparency. To the best of our knowledge, this is the first work attempting to use deep supervision as an auxiliary to build the spatial attention mechanism in hand keypoint detection problem.

\section{ Multi-Scale Deep Supervision Networks}
In this section, we detail the structure of P-MSDSNet. The idea of deep supervision in this paper is represented as a repeated chain of convolutional blocks with spatial attention. The deep supervision is added to multi-scale feature maps to enhance the learning of local and global features. We also explain the learning strategy employed when P-MSDSNet is applied to hand keypoint detection.

\subsection{Deep Supervision Networks with Spatial Attention}
\begin{figure}[ht]
\centering
\includegraphics[width=0.6\textwidth]{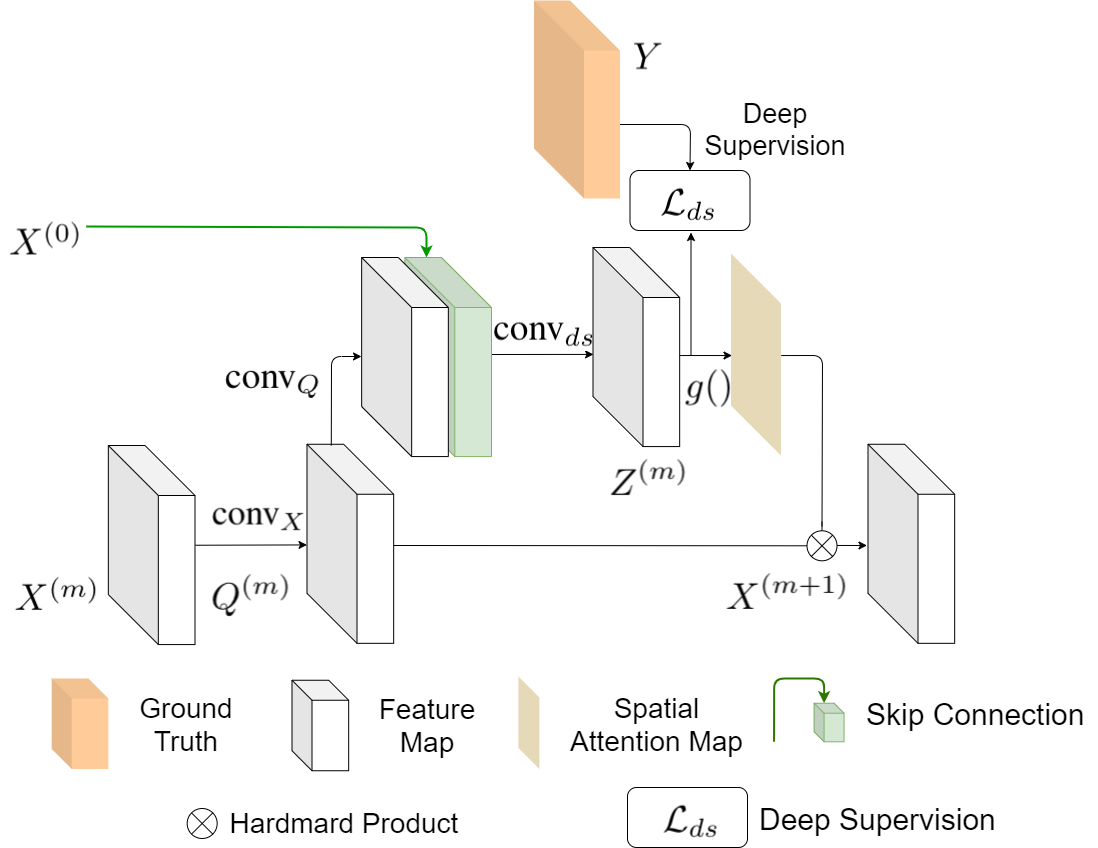}
\caption{An illustration of deep-supervision based spatial attention module. A new branch is added to the feature map and the intermediate output is supervised by the ground truth feature map. The spatial attention map is calculated by summing up the intermediate output feature maps along the channel axis and being activated by the Sigmoid function. Then, the spatial attention feature map is applied on the original feature map through the Hadamard product.}
\label{ds}
\end{figure}    
We will now detail the deep supervision networks with spatial attention. The structure of the network consists of M stages with each stage shown in Figure \ref{ds}. At a stage $m$, the network takes a feature map $X^{(m)}$ as an input and transforms it to an intermediate representation $Q^{(m)}$ by applying a convolutional module $\text{conv}_X(.)$. The module can consist of several convolutional operations, pooling, and normalisation functions. In our implementation, we employed three convolution and batch normalisation operators. The intermediate representation is then used to estimate a spatial attention map and combines with it (through a Hadamard product) to produce the input feature map for the next stage. In contrast to previous work \cite{newell2016stacked} where deep supervision is set up to improve the discrimination of the features, we tailor the deep supervision to adaptively refine the features. To this end, at each stage the model learns the deeply supervised features $Z^{(m)}$ to estimate the spatial attention map. To improve the stability of the gradients during the learning and to fuse different depth features, a skip connection is added to infer the deeply supervised features. The formulation of a stage is shown as follows.
\begin{align}
&Q^{(m)} = \text{conv}_X (X^{(m)})\label{eq:deepsup1}\\
&Z^{(m)} = \text{conv}_{ds}([\text{conv}_Q    (Q^{(m)}),X^{(0)}])\label{eq:deepsup2}\\        
&X^{(m+1)} = g(Z^{(m)})  \otimes Q^{(m)} 
\label{eq:deepsup3}
\end{align}

for $m = 1, .., M$

where $g$ is a spatial attention function \cite{woo2018cbam} and $X^{(1)}=\text{conv}_X(X^{(0)})$ with $X^{(0)} \in \mathbb{R}^{H \times W \times 3}$ is an input image. The final feature map $X^{(M+1)}$, i.e. the output of the M$^{th
}$ layer, is fed into a convolution operation to predict the final output $\hat{Y} = X^{(M+1)} \oplus W^{(f)}$. In \eqref{eq:deepsup1}, \eqref{eq:deepsup2}, \eqref{eq:deepsup3}, $\text{conv}_{X}$, $\text{conv}_{ds}$, $\text{conv}_{Q}$ denote the convolutional blocks.

\subsection{Multi-scale Deep Supervision}
\begin{figure}[ht]
\centering
\includegraphics[width=\textwidth]{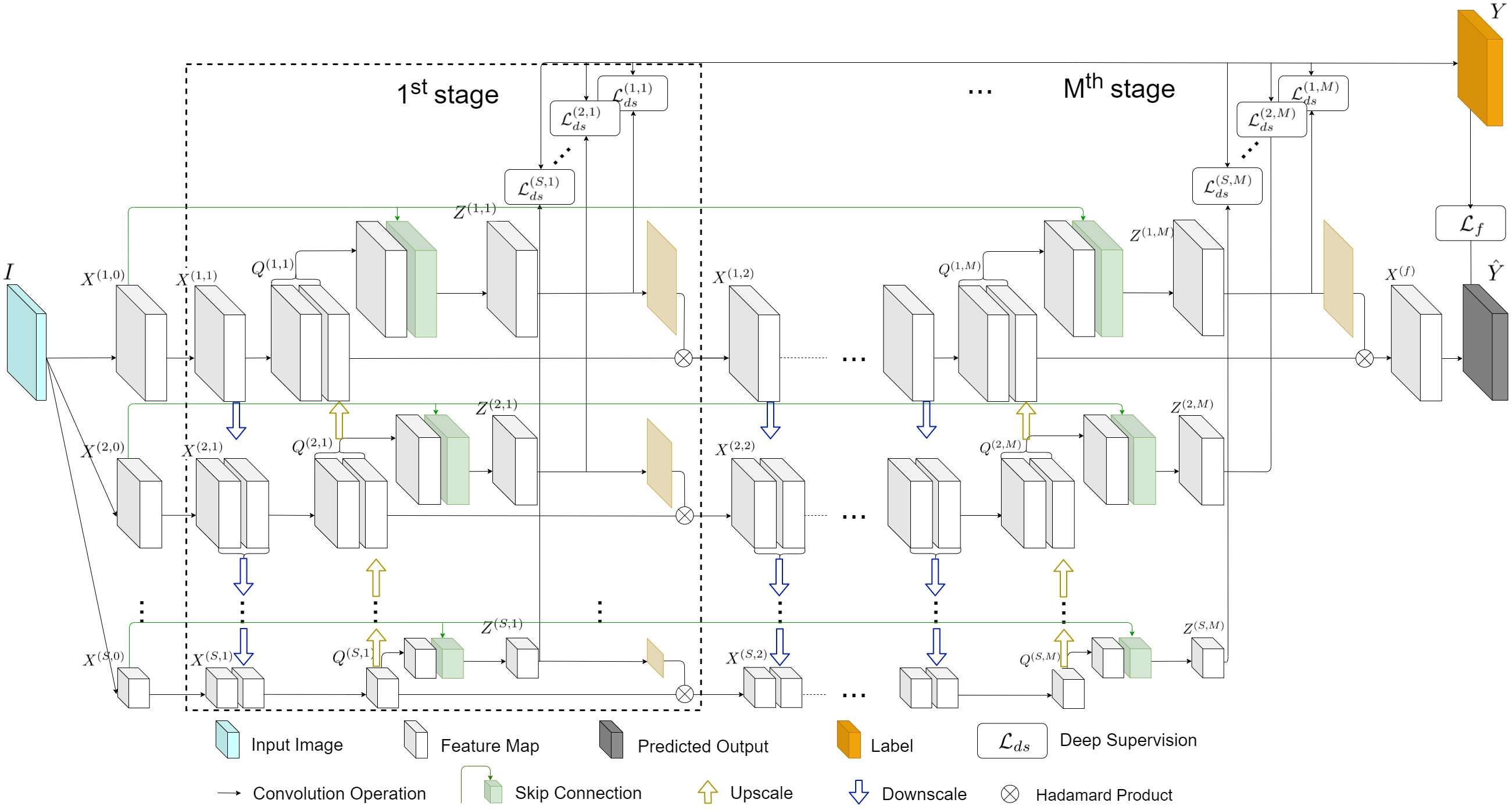}
\caption{An illustration of Multi-Scale Deep Supervision Network. The network fuses different scales' feature maps (upscale and downscale) together at the same depth. The intermediate supervision attention modules are used at different scales (dashed square).}
\label{NS}
\end{figure}
Scale variation poses a critical challenge to the prediction of correct poses \cite{cheng2020higherhrnet,yang2020msb}. To address the issue, we extend P-MSDSNet to learn multi-scale high-level features and propagate them through the depth of the network for the final prediction. P-MSDSNet maintains the effect of deep supervision by adding parallel connections at different scales, meanwhile, stacking the deeply supervised layers at different depth levels.   

A common solution for the scale variance problem is to build a multi-scale feature pyramid \cite{lin2017feature}. These approaches, however, add more computational complexity to the learning and inference of deep CNN models. In \cite{yang2020msb}, a pyramid pooling idea is proposed where a ResNet-based feature map is pooled into a set feature maps with different sizes. However, the employment of ResNet makes it difficult for multi-staging. To improve the modularity, we used convolution with multiple stride sizes to produce feature maps at different scales $\{X^{(s,0)}| s=1, ..., S\}$. This approach allows us to easily incorporate multi-scale features into each stage of the deep supervision network proposed in the previous section (see Figure \ref{NS}). Furthermore, to consolidate context information we fuse the feature maps from multi-scale levels and multi-depth levels. We carefully design the fusion mechanism to keep our network compact. In particular, at a stage $m, (m=1, ..., M)$ we project an input feature map $X^{(s,m)}$ at a scale $s$ to the scale $s-1$ and concatenate it with $X^{(s-1,m)}$. The process starts from $s=1$ and goes downward to the bottom, after that we perform upward fusion, starting from $s=S$. The final architecture of P-MSDSNet is shown in Figure \ref{NS}.

Formally, the inference at each stage is as follows:
\begin{align}
&X^{(s,m)} = [X^{(s,m)},\text{conv}_{down}(X^{(s-1,m)})] \\
&Q^{(s,m)} = [\text{conv}_X(X^{(s,m)}),\text{conv}_{up}(Q^{(s+1,m)})] \\ &Z^{(s,m)} = \text{conv}_{ds}([\text{conv}_Q(Q^{(s,m)}), X^{(s,0)}])\\
& X^{(s,m+1)} = g(Z^{(s,m)}) \otimes Q^{(s,m)}
\end{align}
for $m=1,...,M$ and $s=1, ...,S$.
where $X^{(s,1)}$ is a feature map generated from $X^{(s,0)}$.
\subsection{Learning}
As a multi-stage network, the loss function of P-MSDSNet is a combination of the final loss and the deep supervision losses at all stages. In general, the total loss function is:

\begin{equation}
\label{eq:loss}
    \mathcal{L} = \mathcal{L}_f(\Theta,\theta^{(f)}) + \alpha(\sum_m \beta_m(\sum_s \gamma_s \mathcal{L}_{ds}^{(s,m)}(\Theta,\theta^{(s,m)})))
\end{equation}
 
where $\Theta$ is the parameters of the  convolutional blocks and the attention modules; $\theta^{(s,m)}$ is the parameters of the deep supevision; and $\theta^{(f)}$ is the parameters of the final layers for prediction. In \eqref{eq:loss}, $\alpha$, $\beta_m$ ($m=1, ...,M$) and $\gamma_s$ (s=1, ...,S) are the balance weights; $\alpha$ controls the effect of the deep supervision overall; $\beta_m$ balances the deep supervision at different depth levels while $\gamma_s$ balances the deep supervision at different scale levels. In \cite{lee2015deeply}, a decay function is applied to $\beta_m$ to reduce the effect of the deep supervision gradually during the learning process. In our application of P-MSDSNet to hand keypoint detection, for simplicity we use fixed values for those weights as discussed in the next section. 
\subsection{Implementation for Hand Keypoint Detection}
\subsubsection{Inputs and Labels} The input $I$ of P-MSDSNet for hand keypoint detection is an image of the size $H\times W \times 3$. The label $Y \in \mathbb{R}^{H\times W \times K}$ is a heatmap of $K$ keypoints where each channel  represents the positions of a hand keypoint. To improve the generality, we generate the Gaussian response heatmap with $\boldsymbol{\mu} = \begin{bmatrix}
       x_k  \\
       y_k
     \end{bmatrix}$ and $\boldsymbol{\Sigma} = \begin{bmatrix}
       3 & 3\\
       3 & 3
     \end{bmatrix}$ for each channel, $(x_k,y_k)$ is the position of $k^{th}$ keypoint. The Gaussian heatmaps are the final labels we use for the training.
\subsubsection{P-MSDSNet Architecture.} 
Our P-MSDSNet consists of 3 (or 6) stages ($M=3 or 6$). Five different scales (S=5) are generated using the stride set $\{(1,1), (2,2), (4,4), (8,8), (16,16)\}$. More details of the P-MSDSNet architecture for hand keypoint detection can be found in the Supplementary Material.
\subsubsection{Loss Function.}
Mean Square Error (MSE) is employed to calculate the loss function \eqref{eq:loss}. We set $\beta_m=1$ and $\gamma_s = 1$. In the experiment, we show that good performance can be achieved by setting $\alpha=0.1$. We will also investigate the effect of $\alpha$ in the ablation study.

At each stage, we use the deeply supervised features $Z^{(s,m)}$ to predict the keypoint heatmap at a scale $s$, namely $\hat{Y}^{(s,m)}$. The deep supervision loss $\mathcal{L}_{ds}^{(s,m)} = 
\text{MSE}(\hat{Y}^{(s,m)}  ,Y^{(s)})$ is the mean square error of $\hat{Y}^{(s,m)}$ and $Y^{(s)}$. Here, $Y^{(s)}$ is the resized keypoint heatmap generated by applying bilinear interpolation to $Y$. Since  interpolation approach is approximation, resizing the keypoint heatmap to smaller scales would reduce the effectiveness of deep supervision. Therefore, we only include the deep supervision losses for $s=\{1,2,3\}$ which correspond the the stride sizes $\{(1,1),(2,2),(4,4)\}$.

\section{Experiments}
\subsection{Datasets}
\begin{figure}[h]
\centering
\begin{tabular}{cccccc}
\includegraphics[scale=0.25]{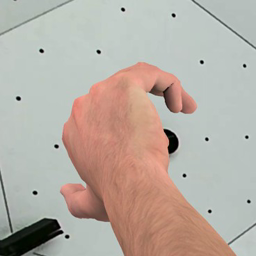}&
\hskip -.3cm
\includegraphics[scale=0.25]{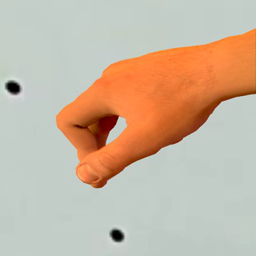}&
\hskip -.3cm
\includegraphics[scale=0.25]{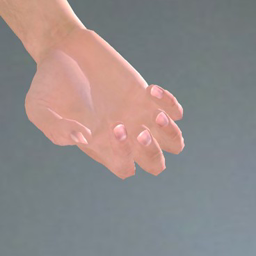}&
\hskip -.3cm
\includegraphics[scale=0.173]{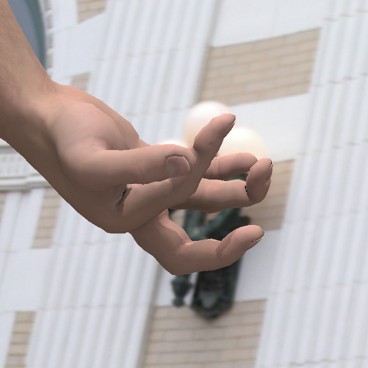}&
\hskip -.3cm
\includegraphics[scale=0.173]{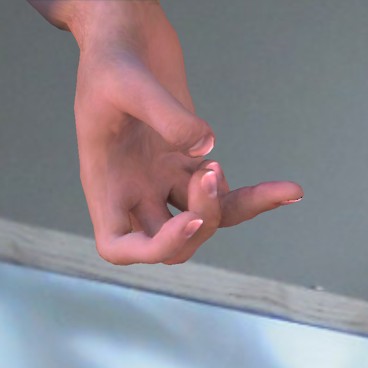}&
\hskip -.3cm
\includegraphics[scale=0.173]{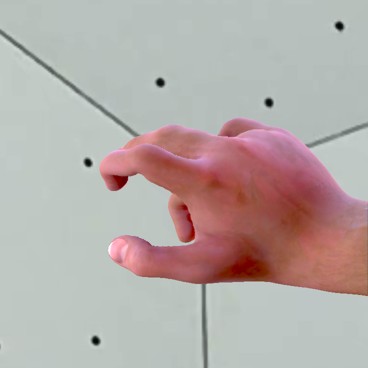}
\hskip -.3cm
\\
\includegraphics[scale=0.25]{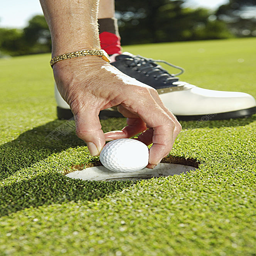} &
\hskip -.3cm
\includegraphics[scale=0.25]{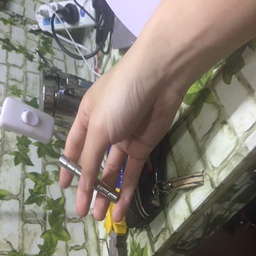} &
\hskip -.3cm
\includegraphics[scale=0.25]{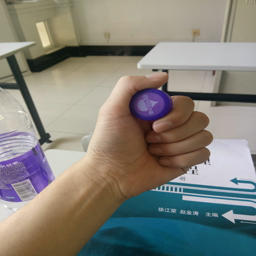} &
\hskip -.3cm
\includegraphics[scale=0.25]{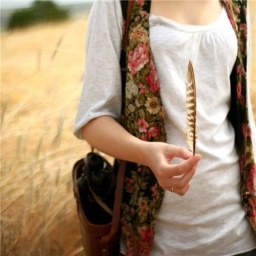} &
\hskip -.3cm
\includegraphics[scale=0.25]{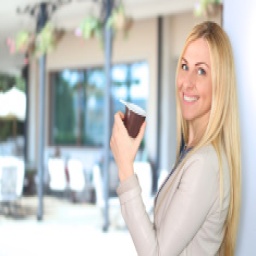} &  
\hskip -.3cm
\includegraphics[scale=0.25]{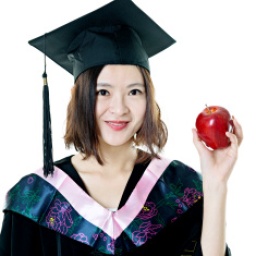}
\hskip -.3cm
\\
\includegraphics[scale=0.25]{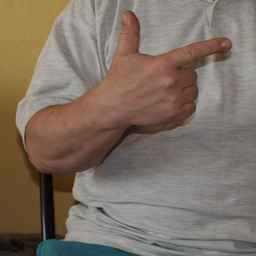}&
\hskip -.3cm
\includegraphics[scale=0.25]{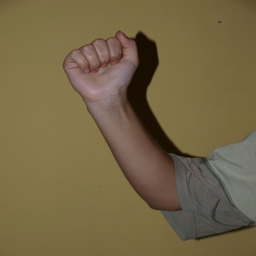}&
\hskip -.3cm
\includegraphics[scale=0.25]{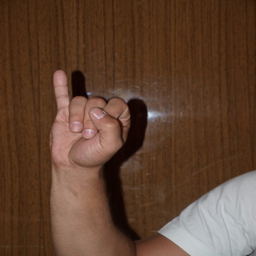}&
\hskip -.3cm
\includegraphics[scale=0.25]{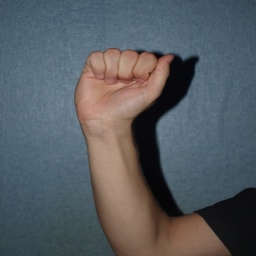}&
\hskip -.3cm
\includegraphics[scale=0.25]{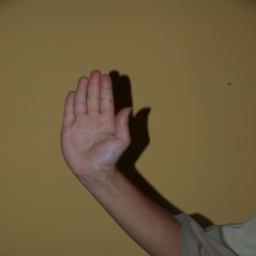}&
\hskip -.3cm
\includegraphics[scale=0.25]{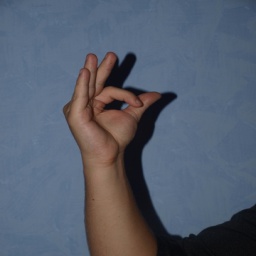}
\hskip -.3cm
\\

\end{tabular}
\vskip -.3cm
\caption{Sample images from CMU Panoptic (Top Row), OneHand10K (Middle Row), and HGR1 (Bottom Row).}
\label{fig:dataset}
\end{figure}

We conducted experiments based on three hand datasets: CMU Panoptic Dataset~\cite{simon2017hand}, Onehand10K Dataset~\cite{wang2018mask} and HGR1 Dataset~\cite{dadashzadeh2019hgr}. The CMU dataset is a synthetic hand dataset that has 7,715 images with sizes varying from $368\times368$ to $1,024\times1,024$. HGR1 dataset has 899 images with sizes varying from $174\times131$ to $640\times480$. Images from both datasets have relatively consistent hand gestures and backgrounds. The OneHand10k dataset has 10,000 images for training and 1,703 images for testing, with random sizes. Images have relatively inconsistent hand gestures and complex backgrounds. Figure \ref{fig:dataset} shows sample images from the datasets.

\subsection{Metrics}
\subsubsection{Percentage of Correct Keypoints (PCK)}
PCK measures the percentage of correct keypoints out of all keypoints. A keypoint is considered correct if the distance (Euclidean distance here) between the predicted and the true keypoint position was within a certain threshold (in pixels).
\begin{align}
&{PCK}_{@t}=\frac{1}{N\times J}\sum_{j=1}^{J}\sum_{n=1}^{N}(||P^{(j)}_{n}-Y^{(j)}_{n}||_{2}<t)
\end{align}
where $N$ is the number of images and $J$ is the number of types of hand keypoints, $P^{(j)}_{n}$ is the predicted position of the $j^{th}$ hand keypoint on the $n^{th}$ image, $Y^{(j)}_{n}$ is the true position of the $j^{th}$ hand keypoint on the $n^{th}$ image and $t$ is the threshold set in advance.

\subsubsection{Mean Per Joint Position Error (MPJPE)}
MPJPE measures the mean of per joint position error for all correctly predicted keypoints based on the threshold.
\begin{align}
&{MPJPE}_{@t}=\frac{1}{\sum_{j=1}^{J}N_{j@t}}\sum_{j=1}^{J}\sum_{n=1}^{N_{j@t}}||P^{(j)}_{n}-Y^{(j)}_{n}||_{2}
\end{align}
where $N_{j@t}$ is the number of images that the $j^{th}$ hand keypoint being correctly detected (based on threshold $t$), $J$ is the number of types of hand keypoints, $P^{(j)}_{n}$ is the predicted position of the $j^{th}$ hand keypoint on the $n^{th}$ image and $Y^{(j)}_{n}$ is the true position of the $j^{th}$ keypoint on the $n^{th}$ image.

\subsection{Results}
An Adam optimizer~\cite{kingma2014adam} was used as the stochastic optimization strategy and learning rate was set to be $5e^{-5}$. Each dataset was partitioned into training, validating and testing sets in a ratio of $8:1:1$. We ran each model several times and report the average performance with standard variance.

For comparison, we used HigherHRNet~\cite{cheng2020higherhrnet}, YOLSE~\cite{wu2017yolse}, MSBFCN~\cite{yang2020msb}, U-Net~\cite{ronneberger2015u} and Stacked Hourglass Model ~\cite{newell2016stacked}. The number of parameters for all models in the experiment is shown in Table \ref{tab:param_num} and this highlights that P-MSDSNet has fewer parameters than other models.

\begin{table}[ht]
\centering
{\scriptsize
\caption{Number of parameters}
\begin{tabular}{l c}
\hline
\textbf{Methods} & \textbf{Number of Parameters} \\
\hline
\hline
\textbf{P-MSDSNet-(3 stages)} & \textbf{2.8M} \\
\textbf{P-MSDSNet-(6 stages)} & \textbf{5M} \\
HigherHRNet	& 3.4M \\
YOLSE & 2.9M \\
MSBFCN & 18.7M \\
U-Net & 4M \\
Hourglass & 8.3M \\
\hline
\end{tabular}
\label{tab:param_num}
}
\end{table}

\begin{figure}[h]
\centering
\begin{tabular}{ccc}
\includegraphics[scale=0.36]{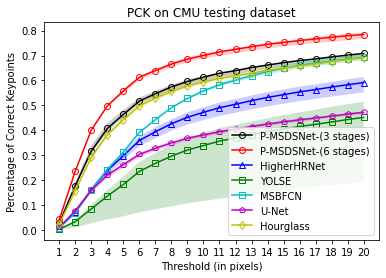}&
\includegraphics[scale=0.36]{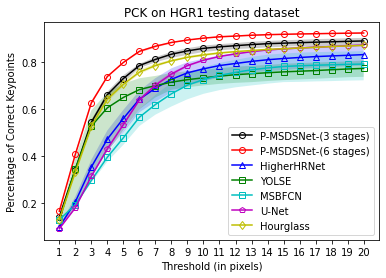}&
\includegraphics[scale=0.36]{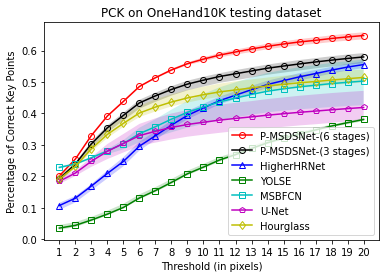}
\end{tabular}
\caption{Different models' Percentage of Correct Keypoints (PCK) on CMU Panoptic Testing Dataset (Left), OneHand10K Dataset (Mid) and HGR1 Testing Dataset (Right).}
\label{fig:pck_comparison}
\end{figure}

Figure \ref{fig:pck_comparison} shows the PCK curves of all six models on three datasets. P-MSDSNet achieved higher performance on all three datasets. Hourglass is the second best model, however, its size is almost 3 times larger than the size of P-MSDSNet. YOLSE has a similar number of parameters as P-MSDSNet but it achieves the worst results among all models.

MPJPE is calculated based on the number of correctly detected keypoints (and this number varies among different methods). Table \ref{tab:mpjpe_comparison} shows the consistent superiority of P-MSDSNet as it achieved low MPJPEs in different datasets. Although MPJPEs for P-MSDSNet are not the lowest in some cases, there are the most correct keypoints (at different thresholds) detected by P-MSDSNet. Although PCK is more popular, for completeness, it is reasonable to use both PCK and MPJPE (with the number of correctly detected keypoints) together for evaluation. From the results for PCK and MPJPE it is evident that P-MSDSNet is a competitive model for hand keypoint detection.
 
\begin{figure}[ht]
\centering
\begin{tabular}{cccccc}
\includegraphics[scale=0.2]{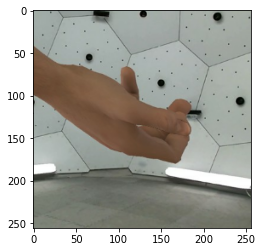} &
  \includegraphics[scale=0.2]{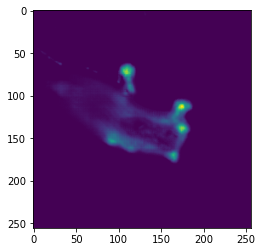}&
  \includegraphics[scale=0.2]{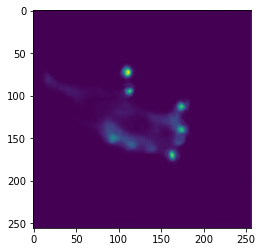}&
  \includegraphics[scale=0.2]{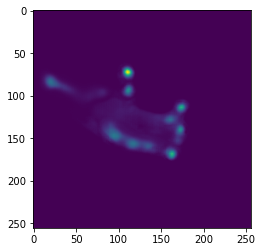}& 
  \includegraphics[scale=0.2]{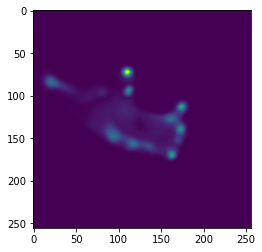} &
  \includegraphics[scale=0.2]{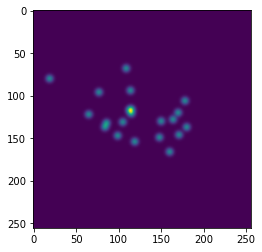}
 \\ 
  \includegraphics[scale=0.2]{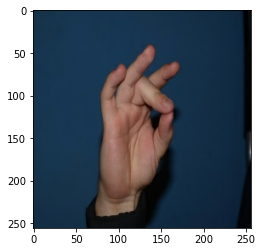}&
  \includegraphics[scale=0.2]{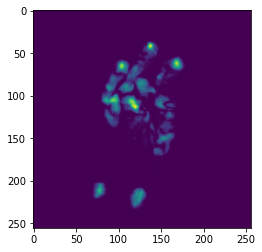}&
  \includegraphics[scale=0.2]{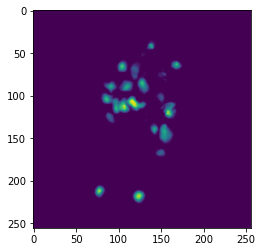}&
  \includegraphics[scale=0.2]{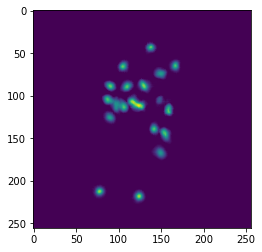}&
  \includegraphics[scale=0.2]{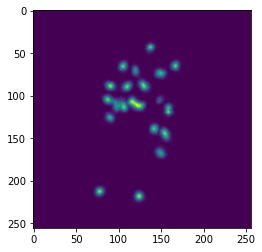}&
  \includegraphics[scale=0.2]{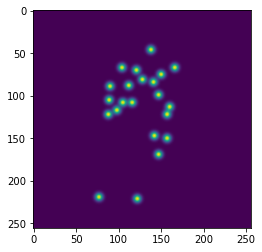}
  \\
  \includegraphics[scale=0.2]{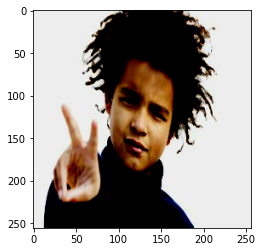}&
  \includegraphics[scale=0.2]{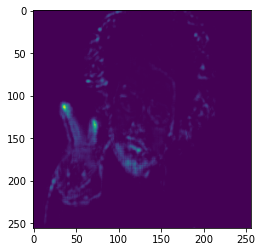}&
  \includegraphics[scale=0.2]{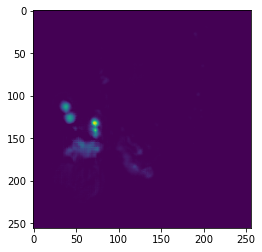}&
  \includegraphics[scale=0.2]{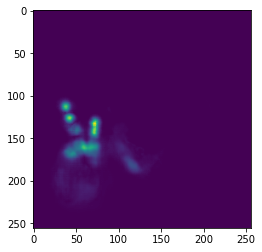}& 
  \includegraphics[scale=0.2]{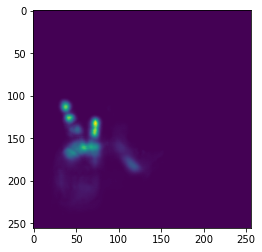}& 
  \includegraphics[scale=0.2]{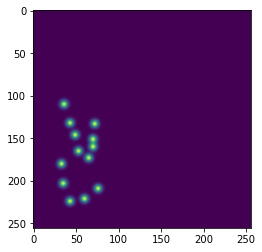}
 \\
Original & Stage 1 & Stage 2 & Stage 3 & Final Prediction & Ground truth
\end{tabular}
\\
\caption{Visualization of P-MSDSNet's different stage deep supervision based spatial attention maps and final predictions}
\label{fig:keypoint_ex}
\end{figure}

\begin{table}[ht]
\centering
{\scriptsize
\caption{Different models' Mean Per Joint Position Error (MPJPE) for correctly predicted keypoints at different thresholds. The average number of correctly detected keypoints (based on a particular threshold) is reported in the parenthesis.}
\begin{tabular}{lccc}
\\
\hline
\textbf{Methods} & \textbf{MPJPE@5pixels} & \textbf{MPJPE@10pixels} & \textbf{MPJPE@20pixels} \\
\hline
\hline
\multicolumn{4}{c}{CMU Testing Dataset}\\
\textbf{P-MSDSNet-(6 stages)} & \textbf{2.19$\pm$0.02 (9,347)} & \textbf{3.06$\pm$0.02 (11,371)} & \textbf{4.21$\pm$0.05 (12,674)} \\
HigherHRNet & 2.78$\pm$0.02 (5,175) & 4.20$\pm$0.02 (7,716) & 6.21$\pm$0.08 (9,606) \\
YOLSE & 3.02$\pm$0.03 (3,261) & 4.77$\pm$0.13 (5,555) & 7.26$\pm$0/61 (7,341) \\
MSBFCN & 2.88$\pm$0.02 (5,492) & 4.58$\pm$0.07 (9,124) & 6.33$\pm$0.12 (11,244) \\
U-Net & 2.58$\pm$0.05 (4,484) & 3.85$\pm$0.09 (6,225) & 5.86$\pm$0.12 (7,666) \\
Hourglass & 2.39$\pm$0.02 (7,483) & 3.44$\pm$0.09 (9,679) & 4.97$\pm$0.03 (11,217) \\
\hline
\multicolumn{4}{c}{OneHand10K Testing Dataset} \\
\textbf{P-MSDSNet-(6 stages)} & \textbf{1.40$\pm$0.07 (16,262)} & \textbf{2.50$\pm$0.03 (20,589)} & \textbf{3.96$\pm$0.07 (23,195)} \\
HigherHRNet & 1.86$\pm$0.06 (9,408) & 3.91$\pm$0.06 (15,097) & 6.39$\pm$0.10 (19,905) \\
YOLSE & 2.27$\pm$0.09 (3,962) & 5.02$\pm$0.10 (8,359) & 8.63$\pm$0.13 (13,654) \\
MSBFCN & 0.89$\pm$0.23 (11,248) & 2.52$\pm$0.50 (15,139) & 4.27$\pm$0.74 (18,002) \\
U-Net & 1.17$\pm$0.22 (11,198) & 2.11$\pm$0.29 (13,352) & 3.48$\pm$3.48 (15,006) \\
Hourglass & 1.40$\pm$0.01 (13,584) & 2.4$\pm$0.01 (16,507) & 3.64$\pm$0.01 (18,419) \\
\hline
\multicolumn{4}{c}{HGR1 Testing Dataset}\\
\textbf{P-MSDSNet-(6 stages)} & \textbf{1.81$\pm$0.03 (3,645)} & \textbf{2.29$\pm$0.05 (4,039)} & \textbf{2.54$\pm$0.04 (4,132)} \\
HigherHRNet & 2.33$\pm$0.03 (2,624) & 3.42$\pm$0.09 (3,457) & 4.12$\pm$0.13 (3,718) \\
YOLSE & 1.79$\pm$0.25 (2,957) & 2.28$\pm$0.48 (3,284) & 2.88$\pm$0.52 (3,458) \\
MSBFCN & 2.25$\pm$0.09 (2,268) & 3.70$\pm$0.12 (3,257) & 4.41$\pm$0.15 (3,531) \\
U-Net & 2.49$\pm$0.02 (2,551) & 3.79$\pm$0.04 (3,628) & 4.46$\pm$0.06 (3,901) \\
Hourglass & 1.95$\pm$0.06 (3,232) & 2.59$\pm$0.08 (3,722) & 3.10$\pm$0.14 (3,896) \\
\hline
\end{tabular}
\label{tab:mpjpe_comparison}
}
\end{table}
Furthermore, in Figure \ref{fig:keypoint_ex} we demonstrate that deep supervision-based spatial attention maps can help guide the network to focus on the propagation of the information around keypoint areas. With its modular structure, i.e. stacking one stage on top of another, P-MSDSNet is able to  refine the attention, step by step. The attention maps offer a means to monitor the learning at each stage. With this mechanism, we can make sure that only the relevant features (around hand areas) can get through for the final prediction step. This idea is inspired by the gating techniques \cite{Hochreiter_1997}.

\section{Ablation Study}
To better understand 1) different ways to fuse multi-scale features, 2) the effectiveness of deep supervision spatial attention module and 3) the balance between final prediction loss and intermediate losses, we present several ablation studies. All experiments in ablation studies were conducted using the same hand datasets in the previous section.

\subsection{Scaling Fusion and Effectiveness of Deep Supervision}
\begin{figure}[h]
\centering
\begin{tabular}{ccc}
\includegraphics[scale=0.36]{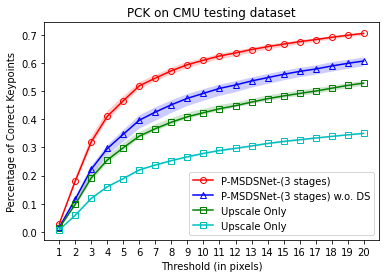}&

\includegraphics[scale=0.36]{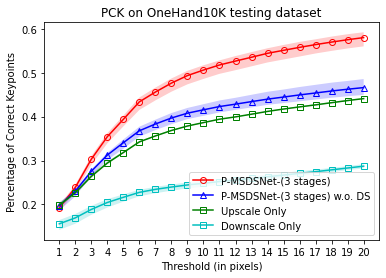}&

\includegraphics[scale=0.36]{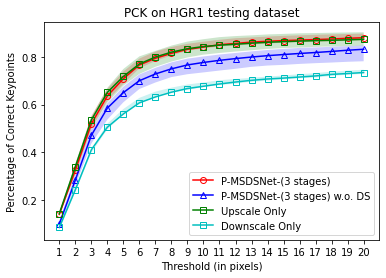}
\end{tabular}
\caption{Ablation study of PCK performance on CMU Panoptic Testing Dataset (Left), OneHand10K Dataset (Mid) and HGR1 Testing Dataset (Right).}
\label{fig:pck_ablation}
\end{figure}

P-MSDSNet has two main novelties: the new design of the upscale-downscale cyclical mechanism to fuse multi-scale features and the deep supervision-based spatial attention module. We conducted the ablation study by changing the multi-scale features fusion mechanism to Upscale Only, Downscale Only, and removing the deep supervision module. Figure \ref{fig:pck_ablation} shows the PCK and Table \ref{tab:ablation} shows the MPJPE.

\begin{table}
\centering
{\scriptsize
\caption{Ablation Study of MPJPE for correctly predicted keypoints at different thresholds on CMU Panoptic Testing Dataset (Left), OneHand10K Dataset (Mid), and HGR1 Testing Dataset (Right). The average number of correctly detected keypoints (based on particular threshold) is reported in the parenthesis.}
\centering
\begin{tabular}{lccc}
\\
\hline
\textbf{Methods} & \textbf{MPJPE@5pixels} & \textbf{MPJPE@10pixels} & \textbf{MPJPE@20pixels} \\
\hline
\hline
\multicolumn{4}{c}{\textbf{CMU Testing Dataset}} \\
\textbf{P-MSDSNet-(6 stages)} & \textbf{2.19$\pm$0.02 (9,347)} & \textbf{3.06$\pm$0.02 (11,371)} & \textbf{4.21$\pm$0.05 (12,674)} \\
P-MSDSNet-(3 stages) & 2.48$\pm$0.02 (7,853) & 3.72$\pm$0.05 (9,933) & 5.49$\pm$0.05 (11,442) \\
HigherHRNet & 2.78$\pm$0.02 (5,175) & 4.20$\pm$0.02 (7,716) & 6.21$\pm$0.08 (9,606) \\
YOLSE & 3.02$\pm$0.03 (3,261) & 4.77$\pm$0.13 (5,555) & 7.26$\pm$0/61 (7,341) \\
MSBFCN & 2.88$\pm$0.02 (5,492) & 4.58$\pm$0.07 (9,124) & 6.33$\pm$0.12 (11,244) \\
U-Net & 2.58$\pm$0.05 (4,484) & 3.85$\pm$0.09 (6,225) & 5.86$\pm$0.12 (7,666) \\
Hourglass & 2.39$\pm$0.02 (7,483) & 3.44$\pm$0.09 (9,679) & 4.97$\pm$0.03 (11,217) \\
\hline
\multicolumn{4}{c}{OneHand10K Testing Dataset} \\
\textbf{P-MSDSNet-(6 stages)} & \textbf{1.40$\pm$0.07 (16,262)} & \textbf{2.50$\pm$0.03 (20,589)} & \textbf{3.96$\pm$0.07 (23,195)} \\
P-MSDSNet-(3 stages) & 1.40$\pm$0.02 (14,567) & 2.56$\pm$0.04 (18,171) & 4.06$\pm$0.08 (20,759) \\
HigherHRNet & 1.86$\pm$0.06 (9,408) & 3.91$\pm$0.06 (15,097) & 6.39$\pm$0.10 (19,905) \\
YOLSE & 2.27$\pm$0.09 (3,962) & 5.02$\pm$0.10 (8,359) & 8.63$\pm$0.13 (13,654) \\
MSBFCN & 0.89$\pm$0.23 (11,248) & 2.52$\pm$0.50 (15,139) & 4.27$\pm$0.74 (18,002) \\
U-Net & 1.17$\pm$0.22 (11,198) & 2.11$\pm$0.29 (13,352) & 3.48$\pm$3.48 (15,006) \\
Hourglass & 1.40$\pm$0.01 (13,584) & 2.4$\pm$0.01 (16,507) & 3.64$\pm$0.01 (18,419) \\
\hline
\multicolumn{4}{c}{HGR1 Testing Dataset}\\
\textbf{P-MSDSNet-(6 stages)} & \textbf{1.81$\pm$0.03 (3,645)} & \textbf{2.29$\pm$0.05 (4,039)} & \textbf{2.54$\pm$0.04 (4,132)} \\
P-MSDSNet-(3 stages) & 1.99$\pm$0.02 (3,278) & 2.64$\pm$0.07 (3,747) & 3.07$\pm$0.10 (3,900) \\
HigherHRNet & 2.33$\pm$0.03 (2,624) & 3.42$\pm$0.09 (3,457) & 4.12$\pm$0.13 (3,718) \\
YOLSE & 1.79$\pm$0.25 (2,957) & 2.28$\pm$0.48 (3,284) & 2.88$\pm$0.52 (3,458) \\
MSBFCN & 2.25$\pm$0.09 (2,268) & 3.70$\pm$0.12 (3,257) & 4.41$\pm$0.15 (3,531) \\
U-Net & 2.49$\pm$0.02 (2,551) & 3.79$\pm$0.04 (3,628) & 4.46$\pm$0.06 (3,901) \\
Hourglass & 1.95$\pm$0.06 (3,232) & 2.59$\pm$0.08 (3,722) & 3.10$\pm$0.14 (3,896) \\
\hline
\\
\\
\\
\end{tabular}
\label{tab:ablation}
}
\end{table}

\begin{figure}[h]
\centering
\begin{tabular}{ccc}
\includegraphics[scale=0.36]{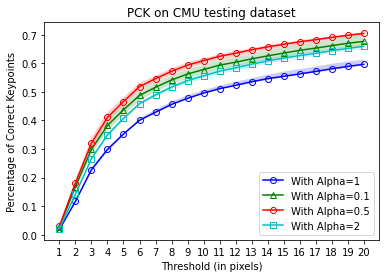}&
\includegraphics[scale=0.36]{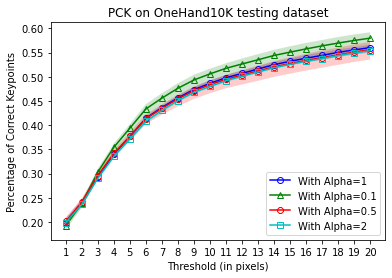}&
\includegraphics[scale=0.36]{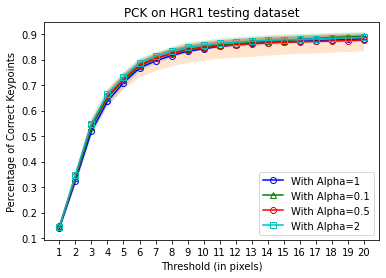}
\end{tabular}
\caption{Balancing weights between intermediate losses and final loss}
\label{fig:pck_alpha}
\end{figure}

These show P-MSDSNet (using the downscale-upscale cyclical pattern for feature fusion) outperforms the one using Upscale Only or Downscale Only. Downscale Only performs worst due to the information flow being spread from large scale (where the prediction is made) to small scale without turning back. This means that the small scale features do not contribute to the final predictions. Compared with Upscale Only, the downscale-upscale cyclical pattern shows a better fusion of multi-scale and different depth information, particularly when facing complex datasets (OneHand10K and CMU).

In addition, P-MSDSNet outperforms the one without deep supervision in terms of PCK and MPJPE, and the gap increases when the dataset becomes more complex. This shows that the deep supervision-based spatial attention module is effective and it improves the hand keypoint detection performance. 

\subsection{Loss Balancing}
We also studied the effect of $\alpha$ value (in (8)) on the model performance. $\alpha$ controls the balancing weights between intermediate losses and the final prediction loss. Figure \ref{fig:pck_alpha} shows the PCK performance and Table \ref{tab:alpha} shows the MPJPE performance for different $\alpha$ values.

\begin{table}
\centering
{\scriptsize
\caption{Effect of $\alpha$ values on the performance.}
\centering
\begin{tabular}{lccc}
\hline
\textbf{Methods} & \textbf{MPJPE@5pixels} & \textbf{MPJPE@10pixels} & \textbf{MPJPE@20pixels} \\
\hline
\hline
\multicolumn{4}{c}{CMU Testing Dataset} \\
$\alpha=1$ & 2.48$\pm$0.02 (5,984) & 3.72$\pm$0.05 (8,137) & 5.49$\pm$0.05 (9,755) \\
$\alpha=0.5$ & 2.30$\pm$0.16 (7,853) & 3.30$\pm$0.10 (9,933) & 4.76$\pm$0.16 (11,442) \\
$\alpha=0.1$ & 2.32$\pm$0.01 (7,380) & 3.36$\pm$0.09 (9,430) & 4.93$\pm$0.12 (10,984) \\
$\alpha=2$ & 2.42$\pm$0.01 (6,898) & 3.55$\pm$0.03 (9,079) & 5.20$\pm$0.03 (10,708) \\
\hline
\multicolumn{4}{c}{OneHand10K Testing Dataset} \\
$\alpha=1$ & 1.40$\pm$0.02 (13,926) & 2.56$\pm$0.04 (17,478) & 4.06$\pm$0.08 (20,055) \\
$\alpha=0.5$ & 1.37$\pm$0.06 (13,937) & 2.51$\pm$0.11 (17,383) & 4.02$\pm$0.17 (19,910) \\
$\alpha=0.1$ & 1.48$\pm$0.01 (14,567) & 2.59$\pm$0.01 (18,171) & 4.05$\pm$0.08 (20,759) \\
$\alpha=2$ & 1.39$\pm$0.01 (13,736) & 2.55$\pm$0.03 (17,250) & 4.08$\pm$0.06 (19,831) \\
\hline
\multicolumn{4}{c}{HGR1 Testing Dataset} \\
$\alpha=1$ & 1.99$\pm$0.02 (3,286) & 2.64$\pm$0.07 (3,801) & 3.07$\pm$0.10 (3,957) \\
$\alpha=0.5$ & 1.93$\pm$0.03 (3,344) & 2.54$\pm$0.06 (3,822) & 2.93$\pm$0.11 (3,961) \\
$\alpha=0.1$ & 1.93$\pm$0.03 (3,278) & 2.54$\pm$0.04 (3,747) & 2.99$\pm$0.08 (3,900) \\
$\alpha=2$ & 1.94$\pm$0.01 (3,356) & 2.55$\pm$0.05 (3,837) & 2.93$\pm$0.04 (3,972) \\
\hline
\end{tabular}
\label{tab:alpha}
}
\end{table}

The allocation for final prediction loss and intermediate losses does affect the model performance. $\alpha=0.1$ works the best on HGR1 and OneHand10K dataset and $\alpha=0.5$ works the best on CMU dataset. Allocating a smaller weights for intermediate losses during training may help improve the P-MSDSNet performance.

\section{Application}
\begin{figure}[h]
\centering
\begin{tabular}{cc}
\includegraphics[scale=0.4]{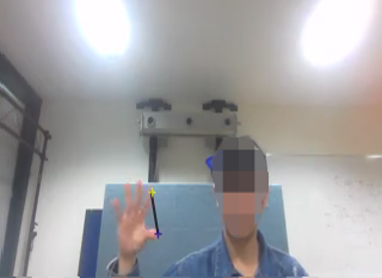}&
    \includegraphics[scale=0.43]{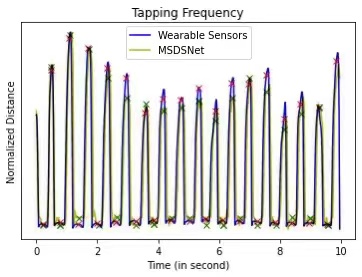}
\\
\end{tabular}
\caption{Application of P-MSDSNet compared to wearable sensors for Neuroscience study. Fingertips detection result (Left) and the finger tapping frequency calculated from the detection of fingertips (Right).}
\label{fig:dataset_neuro}
\end{figure}

In several areas of neuroscience, there is an urgent need to precisely quantify hand movements. For example, people with Parkinson's have slower and less rhythmic patterns of hand movements but there are no accessible methods to quantify these accurately. Currently, finger tapping tests are used to assess progression of disease as well as response to new drugs. This test involves  patients repetitively opening and closing  their index finger and thumb against each other ten times, while a neuroscientist or clinician assesses visually how fast and rhythmically they move, and then applies a subjective score of 0-4 (where 0 is normal and 4 is highly abnormal). Usually wearable sensors have been used to track the finger movements but is inconvenient and not feasible for data collection at a large scale. We apply P-MSDSNet to detect and track hand finger keypoints from a normal laptop webcam (30FPS) in real-time. Before applying P-MSDSNet to estimate finger tapping frequency, we evaluated the model on a dataset of 220 10-second videos of 16 participants doing finger tapping tests. Thumb-tip and index fingertip are the two hand keypoints to be detected. We then apply the trained P-MSDSNet to calculate the finger tapping frequency based on the distance between thumb-tip and index fingertip. Interestingly, it is shown that tracking of P-MSDSNet and that of wearable sensors are consistently similar in Figure \ref{fig:dataset_neuro}. This result implies the reliability of P-MSDSNet for a large scale finger tapping test.

\section{Conclusions and Future work}
We present and apply the P-MSDSNet on the CMU Panoptic~\cite{simon2017hand}, OneHand10K~\cite{wang2018mask} and HGR1 datasets~\cite{dadashzadeh2019hgr}. In particular, our model is better than several comparables in terms of PCK and MPJPE metrics. Additionally, we have also applied P-MSDSNet on a real-life neuroscience domain finger tapping test, which showcased the applicability of  P-MSDSNet.

For the future work, we aim to convert P-MSDSNet to be fully recursive with parameters shared among all stages. We will also develop an adaptive algorithm to determine the optimal depth (number of stages) of the network.

\bibliographystyle{alpha}
\bibliography{reference}

\end{document}